%% file: main.tex
\documentclass[]{spie}  

\usepackage{amsmath,amsfonts,amssymb}
\usepackage{graphicx}
\usepackage{comment}
\usepackage{tikz}
\usetikzlibrary{arrows.meta,positioning,calc,fit,patterns,backgrounds,decorations.pathreplacing}
\usepackage{pgfplots}
\pgfplotsset{compat=1.18}
\usepackage[dvipsnames]{xcolor}
\usepackage{subcaption}
\usepackage{caption}
\usepackage{float}
\usepackage{booktabs}
\usepackage{multirow}
\usepackage{array}
\usepackage{enumitem}
\usepackage{url}
\usepackage{textcomp}
\usepackage[colorlinks=true, allcolors=blue, hypertexnames=false]{hyperref}
\setlist{nosep,leftmargin=*}
\newcommand{\tabhead}[1]{\textbf{#1}}
\newcommand{\CI}{CI}
\newcommand{\ChronoState}{ChronoState}
\newcommand{\Tau}{\tau}
\newcommand{\WSC}{\mathrm{WSC}}
\newcolumntype{L}[1]{>{\raggedright\arraybackslash}p{#1}}

\definecolor{ciBlue}{HTML}{2B6CB0}
\definecolor{ciOrange}{HTML}{DD6B20}
\definecolor{ciGreen}{HTML}{2F855A}
\definecolor{ciRed}{HTML}{C53030}
\definecolor{ciGray}{HTML}{4A5568}
\definecolor{ciLight}{HTML}{F7FAFC}

\title{ChronoState: Hidden Elapsed-Time Conditioning for Temporal-State Action Selection in Frozen-Backbone Language Models}

\author[a]{Sam Siavoshian$^*$}
\author[b]{Omar Ramadan}
\author[b]{Amir Kashif Saeed}
\author[b]{Benjamin A. Johnson}
\author[b]{Amin Mohamed El-Amin Diab}
\author[b]{Benjamin M. Rodriguez}
\affil[a]{Independent Researcher, California, USA}
\affil[b]{Johns Hopkins University, Whiting School of Engineering, Baltimore, MD, USA}
\authorinfo{$^*$Further author information: Send correspondence to Sam Siavoshian.\\
E-mail: samsiavoshian2009@gmail.com}

\begin{document}
\maketitle

\begin{abstract}
\input{sections/abstract}
\end{abstract}


\input{sections/introduction}
\input{sections/background_related_work}
\input{sections/methodology}
\input{sections/results_analysis}
\input{sections/conclusion}

\acknowledgments

The authors thank the Qwen team for releasing the Qwen model family and the developers of open-source tools used in the experiments.
The authors gratefully acknowledge the Computer and Information Technology (CIT) program within the Johns Hopkins University Whiting School of Engineering, Engineering for Professionals, which encompasses Artificial Intelligence, Computer Science, Cybersecurity, Data Analytics, Data Science, and Information Systems Engineering. The support provided through this program was instrumental to this work.

\bibliographystyle{spiebib}
\bibliography{references}

\end{document}

%% file: sections/abstract.tex
Temporal decisions in language-model systems often depend on both symbolic task state and elapsed wall-clock time: caches expire, background jobs complete, quotas reset, deadlines pass, and sessions go stale. We study whether elapsed time can be supplied as a non-token, system-side scalar and composed with visible symbolic state by a frozen-backbone language model. We introduce \ChronoState{}, a compositional temporal-state benchmark in which the prompt carries symbolic state, elapsed seconds $\Tau$ arrive through a hidden chronometric-injection (\CI{}) channel, and the model selects a forced-choice temporal action. Here ``hidden'' means hidden from the user-visible token sequence, not from model computation. With Qwen2.5-3B-Instruct as a frozen bf16 backbone, a 31-dimensional sinusoidal-plus-log time encoding, gated FiLM residual modulation, and a rank-8 LoRA action surface, hidden-time \CI{} reaches $\mathbf{0.9305\pm0.0134}$ accuracy and $\mathbf{0.9410\pm0.0103}$ balanced accuracy on the standard test. No-time and shuffled-time controls fall to $0.5511\pm0.0042$ and $0.3323\pm0.0097$, with high wrong-state consistency under shuffled $\Tau$, supporting causal use of the injected scalar. Generalization remains strong for held-out templates, durations, and multi-constraint compositions, but pure held-out quota-family transfer is weak ($0.5065\pm0.0559$). A fair prompt+LoRA timestamp baseline is stronger ($0.9893\pm0.0052$). Thus \ChronoState{} supports a narrow conclusion: hidden elapsed time can be composed with symbolic task state under direct supervision, without claiming autonomous time tracking, broad unseen-family abstraction, or superiority over prompt timestamps.

%% file: sections/introduction.tex
\section{Introduction}

Many deployed language-model systems interact with external state. A cache may have been filled minutes ago, a background job may have started seconds ago, a quota window may reset tomorrow, or a user may have left an interactive session idle for hours. In these settings, the visible task description alone can be underdetermined: the same symbolic state can require different actions depending on elapsed wall-clock time.

This coupling of symbolic state and elapsed time is especially salient in defense and security workflows. Mission-support and intelligence analysis agents must decide when a track or sensor product remains actionable, when a background fusion or retrieval job has finished, when a quota or rate-limit window has reset, and when an operator session is too stale to continue without re-authentication. In contested or multi-domain environments, writing trusted clock state into the prompt can leak system timing, mix with untrusted text, or be buried in long contexts. A controllable, non-token elapsed-time channel therefore matters not only for general agent reliability but also for secure, time-aware decision support.

The simplest and often strongest approach is to write a timestamp into the prompt. This exposes time to the model through ordinary tokens, allowing the same training and inference interface used for other facts. However, prompt timestamps can be omitted by the system, contradicted by user text, buried in long context, or manipulated if visible time is mixed with untrusted text. Prompt-based agent systems are known to be vulnerable to instruction and tool-selection attacks when untrusted content enters the context~\cite{shi2025toolhijacker}. Prompt timestamps also do not provide a separate continuous control channel that can be independently zeroed, shuffled, or put into conflict with visible text. For controlled research on temporal conditioning, a hidden elapsed-time channel provides a complementary interface.

This paper does not argue that a language model should replace deterministic time-state controllers. For simple cache expiry, quota reset, or deadline checks, symbolic code is more transparent and reliable. The question studied here is narrower and more architectural: when a frozen language-model backbone is used as an action selector, can a small trained conditioning surface cause a system-supplied, non-token elapsed-time scalar to compose with symbolic state in the prompt?

The primary contribution is therefore an evaluation framework for hidden elapsed-time conditioning in frozen-backbone language models. Chronometric injection is the tested interface; \ChronoState{} is the benchmark; the empirical claim is that a hidden scalar can be causally composed with visible symbolic task state under direct supervision. This paper evaluates the interface through scalar interventions rather than re-establishing a full probe and sign-flip mechanistic analysis of chronometric injection. This hierarchy is important because several neighboring ideas already exist, including time-aware language models~\cite{dhingra2022timeaware}, textual and continuous time-aware prompts~\cite{cao2022timeawareprompting}, residual activation steering~\cite{turner2023activation}, and time-aware agent evaluations~\cite{cheng2026temporalblindness}. Our proposal is closest to their intersection, but differs by hiding elapsed time from the token stream and evaluating deterministic temporal actions under scalar interventions.

We study this question using \emph{chronometric injection} (\CI{}): elapsed seconds are encoded as continuous features and injected into the residual stream of a frozen language-model backbone through gated FiLM modulation. \ChronoState{} then asks whether the resulting model can combine hidden elapsed time with visible symbolic state to choose temporal actions. This is different from direct clock readout and different from a single freshness threshold; a temporal action may depend on cache time-to-live, job duration, deadline slack, quota reset interval, session staleness, or a priority rule over several simultaneously active constraints.

The central claim is intentionally narrow. \ChronoState{} does not show autonomous time tracking, subjective time perception, or architectural superiority over prompt timestamps. It shows that, under direct supervision, a hidden elapsed-time channel can support high-accuracy action selection from symbolic temporal state. The same experiments also reveal boundaries: the model follows wrong shuffled hidden time, prompt+LoRA timestamp training is stronger, and transfer to a pure held-out quota family remains weak.

This paper makes four contributions. First, it formulates hidden-time temporal-state reasoning as a composition problem over visible symbolic state $s$ and hidden elapsed seconds $\Tau$. Second, it introduces \ChronoState{}, a forced-choice benchmark covering cache expiration, job completion, deadline feasibility, quota reset, session staleness, and multi-constraint task management. Third, it evaluates chronometric injection against no-time, shuffled-time, prompt-timestamp, LoRA-only, prompt+LoRA, and vanilla controls. Fourth, it characterizes the scope of the result: hidden-time \CI{} solves standard/template/duration/composition splits under direct supervision, but loses to prompt+LoRA and remains weak on pure held-out quota-family transfer.

\begin{figure}[t]
\centering
\resizebox{\textwidth}{!}{%
\begin{tikzpicture}[font=\sffamily\scriptsize,>=Latex,line width=0.48pt]
\tikzset{
  panel/.style={draw=black!35, rounded corners=2mm, fill=black!2, inner sep=0pt},
  statebox/.style={draw=ciBlue!60, rounded corners=1.4mm, fill=ciBlue!8, align=left, inner sep=4pt},
  timebox/.style={draw=ciOrange!70, rounded corners=1.4mm, fill=ciOrange!10, align=center, inner sep=4pt},
  actionbox/.style={draw=ciGreen!70, rounded corners=1.4mm, fill=ciGreen!10, align=center, inner sep=4pt},
  module/.style={draw=black!45, rounded corners=1.4mm, fill=white, align=center, inner sep=4pt},
  train/.style={draw=ciGreen!70, rounded corners=1.4mm, fill=ciGreen!10, align=center, inner sep=4pt},
  hiddenarrow/.style={ciOrange!90, thick, dashed, -{Latex[length=2mm]}},
  visar/.style={ciBlue!85, thick, -{Latex[length=2mm]}}
}

\node[panel, minimum width=7.15cm, minimum height=4.35cm, anchor=north west] (A) at (0,0) {};
\node[anchor=north west,font=\bfseries] at ([xshift=0.18cm,yshift=-0.18cm]A.north west) {(A) Problem};
\node[statebox, text width=6.15cm, anchor=north] (promptA) at ([yshift=-0.72cm]A.north) {\textbf{Visible symbolic state}\\ Cache was filled earlier; TTL = 30 min.\\Choose: \texttt{REUSE} / \texttt{WAIT} / \texttt{REFRESH} / \texttt{RETRIEVE}};
\node[timebox, text width=1.62cm, anchor=west] (tauA1) at ([xshift=0.55cm,yshift=-2.55cm]A.north west) {hidden\\$\Tau=5$ min};
\node[actionbox, text width=1.72cm, anchor=west] (actA1) at ([xshift=4.65cm,yshift=-2.55cm]A.north west) {\texttt{REUSE}};
\draw[hiddenarrow] (tauA1) -- (actA1);
\node[timebox, text width=1.62cm, below=0.42cm of tauA1] (tauA2) {hidden\\$\Tau=2$ hr};
\node[actionbox, text width=1.72cm, anchor=west] (actA2) at ([xshift=4.65cm,yshift=-3.45cm]A.north west) {\texttt{REFRESH}};
\draw[hiddenarrow] (tauA2) -- (actA2);

\node[panel, minimum width=10.15cm, minimum height=4.35cm, anchor=north west] (B) at (7.55,0) {};
\node[anchor=north west,font=\bfseries] at ([xshift=0.18cm,yshift=-0.18cm]B.north west) {(B) Chronometric-injection interface};
\node[statebox, text width=1.55cm] (tokens) at ([xshift=1.05cm,yshift=-1.25cm]B.north west) {visible\\tokens $x$};
\node[module, text width=2.05cm, right=0.85cm of tokens] (backbone) {frozen LM\\backbone\\$f_{\theta}$};
\node[train, text width=2.15cm, right=0.85cm of backbone] (lora) {LoRA action\\surface\\$\rightarrow$ A/B/C/D};
\draw[visar] (tokens) -- (backbone);
\draw[visar] (backbone) -- (lora);
\node[timebox, text width=0.95cm] (tauB) at ([xshift=1.10cm,yshift=-3.10cm]B.north west) {$\Tau$};
\node[module, text width=1.70cm, right=0.55cm of tauB] (enc) {$\chi(\Tau)$\\sin/cos/log};
\node[train, text width=3.25cm, right=0.60cm of enc] (film) {gated FiLM\\$h'_{\ell}=h_{\ell}+\alpha_{\ell}\odot(\gamma_{\ell}\odot h_{\ell}+\beta_{\ell})$};
\draw[hiddenarrow] (tauB) -- (enc);
\draw[hiddenarrow] (enc) -- (film);
\draw[hiddenarrow] (film.north) to[out=90,in=-90] (backbone.south);
\node[anchor=south, align=center, text=black!70] at ([yshift=0.16cm]B.south) {hidden = non-token system-side scalar, not latent time perception};

\node[panel, minimum width=17.70cm, minimum height=2.65cm, anchor=north west] (C) at (0,-4.75) {};
\node[anchor=north west,font=\bfseries] at ([xshift=0.18cm,yshift=-0.18cm]C.north west) {(C) Scalar interventions};
\def\barw{5.6}
\newcommand{\ciinterventionbar}[5]{%
  \node[anchor=east] at ([xshift=3.05cm,yshift=#4]C.north west) {#1};%
  \pgfmathsetmacro{\len}{#2*\barw}%
  \draw[fill=#5!58,draw=#5!85] ([xshift=3.25cm,yshift=#4-0.13cm]C.north west) rectangle ++(\len cm,0.26cm);%
  \node[anchor=west] at ([xshift=9.05cm,yshift=#4]C.north west) {#3};%
}
\ciinterventionbar{true hidden $\Tau$}{0.931}{0.931}{-0.82cm}{ciGreen}
\ciinterventionbar{zero $\Tau$}{0.551}{0.551}{-1.38cm}{ciGray}
\ciinterventionbar{shuffled $\Tau$}{0.332}{0.332}{-1.94cm}{ciRed}
\node[draw=ciRed!70, fill=ciRed!8, rounded corners=1.4mm, text width=6.0cm, align=center, anchor=east] at ([xshift=-0.35cm,yshift=-1.36cm]C.east) {Shuffling $\Tau$ collapses original-label accuracy while wrong-state consistency remains high.};
\end{tikzpicture}%
}
\caption{Overview of \ChronoState{} and chronometric injection. (A) The same visible symbolic state can require different actions under different hidden elapsed times. (B) The elapsed-time scalar is encoded outside the prompt and injected through gated FiLM modulation into a frozen-backbone model with a small LoRA action surface. (C) Zeroing or shuffling the scalar tests whether predictions causally depend on the hidden time channel.}
\label{fig:overview}
\end{figure}
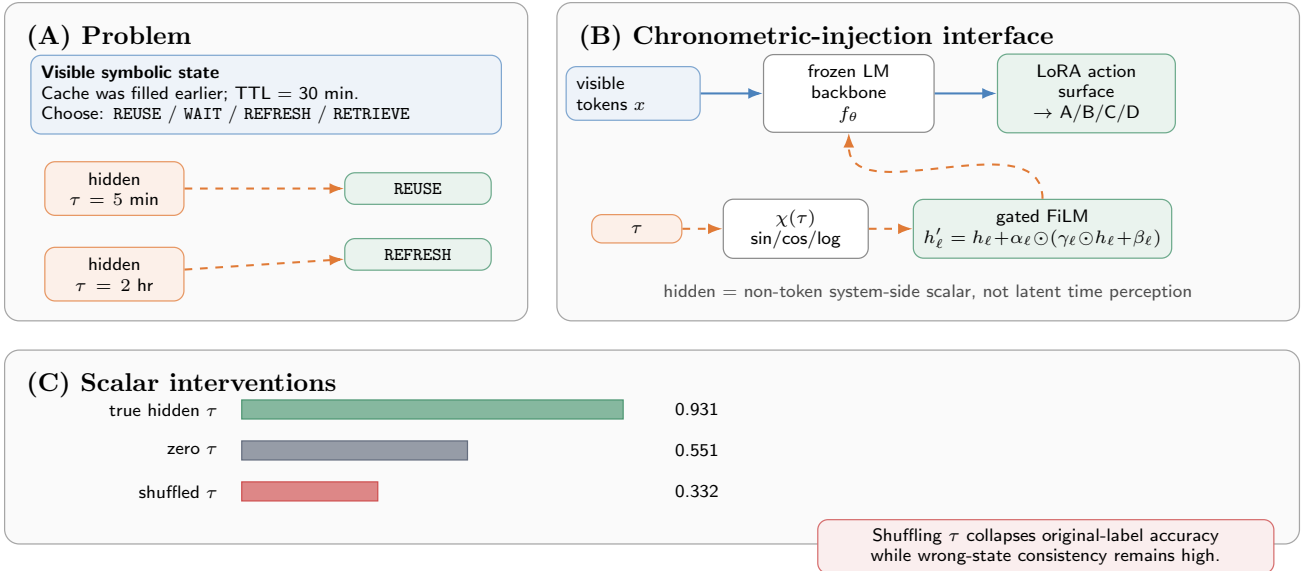

Fig.~\ref{fig:overview} summarizes the end-to-end setup: symbolic state is visible in the prompt, elapsed time is supplied through a hidden scalar channel, and scalar interventions test whether the model actually uses the supplied time.

%% file: sections/background_related_work.tex
\section{Background and Related Work}

\subsection{Time-aware language models and prompts}

Standard Transformer positional encodings represent token order, but token position is not elapsed wall-clock time~\cite{vaswani2017attention}. Time-aware language modeling has been studied primarily as a way to handle facts whose truth changes over calendar time. Dhingra et al. train language models jointly with timestamps and introduce temporal knowledge probes for facts that vary over time~\cite{dhingra2022timeaware}. Cao and Wang study time-aware prompting for generation, comparing natural-language timestamp prompts with linear prompts that convert timestamps into continuous vectors~\cite{cao2022timeawareprompting}. These works are methodologically close because they inject temporal information into language-model computation, but their timestamps index documents or knowledge states. \ChronoState{} instead supplies elapsed seconds since a system-defined reference event and evaluates action selection over symbolic temporal predicates.

A 2025 preprint on time-injected LLMs proposes periodically supplying an LLM with standardized time prompts to elicit agentic behavior without architectural modification or retraining~\cite{eldridge2025timeinjected}. That work is close in motivation because it treats time as an external signal for agent behavior. Our setting differs because elapsed time is not written into the visible prompt, the model is trained with a residual conditioning path, and evaluation uses deterministic state--time actions plus scalar ablations.

\subsection{Temporal reasoning and temporal agents}

Temporal reasoning benchmarks such as TRAM and Test of Time evaluate LLMs on event ordering, temporal arithmetic, duration, frequency, and related natural-language reasoning tasks~\cite{wang2024tram,fatemi2024testoftime}. These benchmarks are broader than \ChronoState{} in their linguistic and temporal-reasoning coverage. \ChronoState{} is narrower: it isolates elapsed wall-clock time as a system variable and asks whether it composes with explicit symbolic state.

Agent work also motivates the problem. Generative agents use memory, reflection, planning, and time-stamped experiences to produce temporally coherent behavior over simulated days~\cite{park2023generativeagents}. More directly, recent work on temporally blind LLM agents studies failures that occur when agents ignore real time elapsed between turns and therefore overuse stale context or unnecessarily repeat tool calls~\cite{cheng2026temporalblindness}. \ChronoState{} focuses on the same broad problem class---deciding whether old state remains usable---but removes the full agent loop and studies a controlled hidden-state interface. The same failure modes appear in real-time sensing and autonomous systems pipelines, where track currency, sensor revisit intervals, and tool-call cadence depend on trusted elapsed time rather than on token order alone; a hidden scalar channel is a natural interface when that clock state should condition routing without entering the visible prompt.

\subsection{Side-channel conditioning and activation interventions}

Feature-wise linear modulation conditions intermediate activations through learned scale and shift transformations~\cite{perez2018film}. Adaptive layer-normalization and zero-initialized conditioning mechanisms are widely used to insert side information while initially preserving a pretrained model's behavior~\cite{peebles2023dit}. Soft prompts, prefix tuning, prompt tuning, and adapters provide related ways to condition a frozen model through trainable non-token or quasi-token parameters~\cite{houlsby2019adapters,li2021prefix,lester2021prompt}. \ChronoState{} uses this family of mechanisms as a hidden elapsed-time input interface.

Residual-stream steering is also related. Activation Addition modifies internal activations during inference by adding steering vectors computed from contrastive prompts~\cite{turner2023activation}. Chronometric injection differs in that the intervention is a structured scalar with known external semantics, passed through a trained conditioning module and evaluated against a deterministic temporal-state oracle.

\subsection{Parameter-efficient adaptation}

Low-rank adaptation modifies a small number of trainable parameters while keeping most pretrained weights fixed~\cite{hu2021lora}. This is useful for studying controlled channels because the base model remains frozen. In \ChronoState{}, LoRA supplies an action-selection surface over chrono-modulated hidden states. LoRA-only and prompt+LoRA baselines distinguish the hidden-time channel from ordinary parameter-efficient adaptation.

\subsection{Mechanistic and behavioral analysis}

Mechanistic studies show that language models can contain linearly recoverable representations of structured variables, including spatial and temporal concepts~\cite{gurnee2023language}. \ChronoState{} is complementary: it does not only ask whether time is represented, but whether an externally supplied hidden time scalar is used to recompute temporal predicates and actions when combined with visible symbolic state. Table~\ref{tab:positioning} summarizes how this positioning differs from neighboring lines of work.

\begin{table}[t]
\centering
\caption{Positioning relative to nearby work. The proposed combination is not that time or activation conditioning is new in isolation; the distinct object of study is hidden elapsed wall-clock time as a system-supplied scalar side channel for symbolic temporal-action selection.}
\label{tab:positioning}
\small
\begin{tabular}{@{}p{0.20\textwidth}p{0.31\textwidth}p{0.40\textwidth}@{}}
\toprule
\tabhead{Neighboring line} & \tabhead{Shared idea} & \tabhead{Difference in \ChronoState{}}\\
\midrule
Time-aware LMs & Timestamps condition language models over changing facts~\cite{dhingra2022timeaware}. & We use elapsed seconds since a reference event, not document/calendar time, and score temporal actions rather than temporal factual recall.\\
Time-aware prompting & Textual and continuous timestamp prompts improve generation~\cite{cao2022timeawareprompting}. & Our scalar is hidden from the token stream and can be causally shuffled, zeroed, or put into conflict with visible text.\\
Time-injected LLM preprint & Periodic time prompts are used to elicit agent behavior~\cite{eldridge2025timeinjected}. & We use residual-stream conditioning and supervised forced-choice temporal-state actions, not prompt-only time signals.\\
Temporal reasoning benchmarks & Benchmarks test order, arithmetic, duration, and related reasoning~\cite{wang2024tram,fatemi2024testoftime}. & We isolate system-side elapsed time plus symbolic state rather than broad temporal QA.\\
Temporal agent evaluations & Agents can ignore elapsed real time between turns~\cite{cheng2026temporalblindness}. & We remove the agent loop and directly test a hidden-time conditioning interface.\\
Activation steering & Internal activations can be modified to steer model outputs~\cite{turner2023activation}. & Our intervention is a known scalar state variable with a deterministic oracle, not a contrastive behavioral direction.\\
\bottomrule
\end{tabular}
\end{table}

%% file: sections/methodology.tex
\section{Methodology}

\subsection{Scope and design rationale}
\label{sec:positioning}

This subsection states several choices that otherwise can be misread as stronger claims than intended. The contribution is not that sinusoidal time encodings, FiLM, LoRA, activation interventions, or timestamp conditioning are new in isolation. The contribution is the controlled combination of a system-supplied elapsed-time scalar, hidden from the user-visible token stream, with symbolic temporal-action evaluation and causal scalar interventions.

\subsubsection{Hidden-time interface}

``Hidden time'' means that elapsed time is not represented as user-visible text. It is supplied as a system-side scalar and injected into model activations. It is hidden from the token sequence, from ordinary prompt manipulation, and from the model's text parser. It is not hidden from model computation: the model explicitly receives $\Tau$ through the conditioning path. We assume the scalar is supplied by a trusted system component; validating that source is outside the model and remains a deployment requirement. More precise names would be \emph{non-token elapsed-time conditioning} or \emph{system-side scalar time conditioning}; we retain \CI{} as shorthand for the implemented residual-stream interface.

\subsubsection{Contribution scope}

The main contribution is the benchmarked evaluation setup, not any single architectural component. \ChronoState{} asks whether a frozen-backbone model can use an externally supplied elapsed-time scalar when selecting actions from visible symbolic task state. Chronometric injection is one tested implementation of this interface. The scalar interventions (zeroing, shuffling, and prompt conflict) are central because they distinguish causal use of the supplied time from memorization of templates or action priors.

\subsubsection{Role of deterministic controllers}

For many production decisions, code is the right answer. If the only required operation is ``refresh if $\Tau\geq$ TTL,'' then a controller should compute that predicate outside the model. \ChronoState{} is not a proposal to replace symbolic controllers. It is a probe of whether model behavior can be conditioned on system-supplied continuous state when that state must influence language-model action selection, routing, or explanation. This distinction is important for deployment: a safe system can use symbolic validation around a learned hidden-time-conditioned model rather than delegating all time logic to the model.

\subsubsection{Relation to prompt timestamps}

The fair prompt+LoRA timestamp baseline outperforms hidden-time \CI{} in the reported experiments. The motivation for hidden time is therefore not raw accuracy superiority on \ChronoState{}. Instead, hidden time provides a separate system-side control channel. It can be zeroed, shuffled, contradicted by visible text, or validated independently of the prompt. This makes hidden-time conditioning useful as a research object for studying system-state interfaces, even when prompt timestamps remain the stronger supervised baseline.

\subsubsection{Definition of composition}

We use \emph{composition} in a limited benchmark sense. The correct action is not a function of elapsed time alone and not a function of visible prompt state alone. It is a deterministic predicate over both. In multi-constraint items, several predicates may be active at once, and the oracle applies a fixed priority policy. Thus composition means state--time predicate evaluation plus deterministic conflict resolution, not open-ended temporal reasoning over arbitrary world knowledge.

\subsubsection{Methodological rationale}

The method combines four pragmatic choices.
\begin{enumerate}[label=(D\arabic*)]
    \item \textbf{A hidden scalar rather than a timestamp token.} This allows the scalar to be zeroed, shuffled, or placed in conflict with visible text while holding the prompt nearly fixed.
    \item \textbf{FiLM-style residual modulation.} FiLM provides a lightweight affine conditioning mechanism and can be zero-initialized so that the pretrained backbone is initially unchanged~\cite{perez2018film,peebles2023dit}.
    \item \textbf{A small LoRA action surface.} LoRA supplies trainable task adaptation without updating the backbone~\cite{hu2021lora}. The method is therefore a frozen-backbone adaptation, not a claim about a completely frozen model with no trainable surface.
    \item \textbf{A fixed sinusoidal-plus-log time encoding.} This representation is inspired by continuous time-feature maps such as Time2Vec~\cite{kazemi2019time2vec}, but it is an engineering choice. Simpler raw, log-only, learned, and Time2Vec-style encodings remain necessary ablations.
\end{enumerate}

These choices are not claimed to be uniquely optimal. Prefix tuning, prompt tuning, adapters, scalar embeddings, and activation engineering are natural alternatives~\cite{houlsby2019adapters,li2021prefix,lester2021prompt,turner2023activation}. The claim evaluated here is that this particular hidden scalar interface is viable and causally controllable on \ChronoState{}.

\subsection{Chronometric injection}

This subsection details the chronometric injection interface: how elapsed seconds are encoded, how they modulate the residual stream of a frozen language-model backbone through gated FiLM, and how zero initialization preserves the pretrained model at the start of training.

\subsubsection{Problem setting}

Let $f_\theta$ be an autoregressive language-model backbone with frozen parameters $\theta$. Let $x$ denote the visible prompt, $s$ denote symbolic state described in the prompt, and $\Tau\in\mathbb{R}_{\geq0}$ denote elapsed monotonic seconds since a system-defined reference event. Examples of reference events include cache fill time, job start time, last successful retrieval, quota-window start, or last user interaction.

\ChronoState{} studies a conditioned model
\begin{equation}
    g_{\theta,\phi}(x,s,\Tau),
\end{equation}
where $\phi$ denotes the small trained conditioning and action-selection surface. The backbone weights $\theta$ are frozen; the complete system is therefore a frozen-backbone model, not a fully frozen model.

The model does not measure time internally. It receives $\Tau$ from an external clock or controller at inference time. The question is whether this scalar can be combined with visible state variables in the hidden computation of a frozen-backbone LLM. Throughout this paper, ``hidden'' means hidden from the user-visible token sequence and from ordinary prompt editing. It does not mean that $\Tau$ is latent or inaccessible to the model.

\subsubsection{Elapsed-time encoding}

Elapsed seconds are mapped into a 31-dimensional fixed feature vector:
\begin{equation}
\begin{aligned}
\chi(\Tau)=\big[&\sin(\Tau/T_1),\cos(\Tau/T_1),\ldots,\\
&\sin(\Tau/T_{15}),\cos(\Tau/T_{15}),\log(1+\Tau)\big]^\top .
\end{aligned}
\end{equation}
The divisors are
\begin{equation}
\begin{aligned}
T\in\{&2,4,8,16,32,64,128,256,512,1024,\\
&4096,16384,65536,86400,604800\}\ \mathrm{s}.
\end{aligned}
\end{equation}
The implementation uses $\sin(\Tau/T_k)$ and $\cos(\Tau/T_k)$ directly, without a $2\pi$ multiplier. Thus $T_k$ is a radian-scale divisor and the mathematical sinusoid period is $2\pi T_k$. The logarithmic feature provides a monotone nonperiodic coordinate. We use this encoding as a fixed interface rather than a tuned architecture; ablations over raw, log-only, sinusoid-only, and learned scalar encodings remain future work.

\subsubsection{Residual-stream modulation}

At transformer layer $\ell$, let $h_\ell\in\mathbb{R}^d$ be the residual-stream hidden state. Chronometric injection computes
\begin{align}
\gamma_\ell &= W^\gamma_\ell\chi(\Tau)+b^\gamma_\ell,\\
\beta_\ell &= W^\beta_\ell\chi(\Tau)+b^\beta_\ell,
\end{align}
and applies gated FiLM modulation:
\begin{equation}
 h'_\ell=h_\ell+\alpha_\ell\odot\left(\gamma_\ell\odot h_\ell+\beta_\ell\right),
\label{eq:film}
\end{equation}
where $\alpha_\ell\in\mathbb{R}^d$ is a learned per-channel gate. Trainable parameters are confined to chrono projectors, gates, and the specified LoRA action surface. This preserves the pretrained backbone while allowing supervised adaptation to the temporal-state task.

\subsubsection{Zero initialization}

The projectors use a zero-initialized conditioning scheme:
\begin{equation}
\alpha_\ell=0,\quad W^\gamma_\ell=0,\quad b^\gamma_\ell=\mathbf{1},\quad W^\beta_\ell=0,\quad b^\beta_\ell=0.
\end{equation}
At initialization, Eq.~\ref{eq:film} reduces to $h'_\ell=h_\ell$, so the model initially behaves exactly like the frozen backbone. The gate still receives nonzero gradient because
\begin{equation}
\frac{\partial h'_\ell}{\partial \alpha_\ell}\bigg|_{\mathrm{init}}=h_\ell.
\end{equation}
The chrono projectors receive time-dependent gradients once the gate departs from zero. This initialization is intended to avoid an abrupt distribution shift in the residual stream at the start of training.

\subsection{\ChronoState{} benchmark}

This subsection defines the \ChronoState{} task, the deterministic oracle decision policy, the six temporal families, and the train and held-out evaluation splits used throughout the experiments.

\subsubsection{Task definition}

Each \ChronoState{} record contains a visible prompt $x_i$, symbolic state $s_i$, hidden elapsed time $\Tau_i$, deterministic temporal predicates $z_i$, a gold action $y_i$, and split metadata. In the hidden-time condition, $\Tau_i$ is supplied only through the chrono channel and is not written into the visible prompt.

The task is forced-choice action selection over the finite action set
\begin{equation}
\mathcal{A} = \{\texttt{REUSE},\ \texttt{WAIT},\ \texttt{RETRIEVE},\ \texttt{REFRESH},\ 
\texttt{WAIT\_QUOTA},\ \texttt{DEADLINE\_MISSED},\ \texttt{ASK\_CONFIRM}\}.
\end{equation}
Four A/B/C/D choices are sampled for each item, always including the gold action $y_i$. Answer letters are balanced by seed. Predictions are scored by the probability assigned to the next answer letter and then mapped back to the corresponding action in $\mathcal{A}$.

\subsubsection{Decision policy}
\label{sec:decision_policy}

\ChronoState{} uses a deterministic oracle policy $\pi(s,\Tau)$ to map symbolic state and elapsed time to a gold action. The single-family policies are:
\begin{align}
\pi_{\mathrm{cache}}(T_{\mathrm{ttl}},\Tau)&=
\begin{cases}
\texttt{REUSE}, & \Tau<T_{\mathrm{ttl}},\\
\texttt{REFRESH}, & \Tau\geq T_{\mathrm{ttl}},
\end{cases}\\
\pi_{\mathrm{job}}(T_{\mathrm{job}},\Tau)&=
\begin{cases}
\texttt{WAIT}, & \Tau<T_{\mathrm{job}},\\
\texttt{RETRIEVE}, & \Tau\geq T_{\mathrm{job}},
\end{cases}\\
\pi_{\mathrm{deadline}}(D,R,\Tau)&=
\begin{cases}
\texttt{WAIT}, & D-\Tau\geq R,\\
\texttt{DEADLINE\_MISSED}, & D-\Tau<R,
\end{cases}\\
\pi_{\mathrm{quota}}(T_{\mathrm{reset}},\Tau)&=
\begin{cases}
\texttt{WAIT\_QUOTA}, & \Tau<T_{\mathrm{reset}},\\
\texttt{RETRIEVE}, & \Tau\geq T_{\mathrm{reset}},
\end{cases}\\
\pi_{\mathrm{stale}}(T_{\mathrm{stale}},\Tau)&=
\begin{cases}
\texttt{REUSE}, & \Tau<T_{\mathrm{stale}},\\
\texttt{ASK\_CONFIRM}, & \Tau\geq T_{\mathrm{stale}}.
\end{cases}
\end{align}
Here $T_{\mathrm{ttl}}$ is cache time-to-live, $T_{\mathrm{job}}$ is expected job duration, $D$ is visible time until deadline at the reference point, $R$ is required completion time, $T_{\mathrm{reset}}$ is the quota reset interval, and $T_{\mathrm{stale}}$ is the staleness threshold.

For multi-constraint examples, the oracle evaluates active predicates in a fixed priority order. Inactive predicates are skipped. The policy is:
\begin{enumerate}
    \item if an active deadline predicate is infeasible, output \texttt{DEADLINE\_MISSED};
    \item else, if an active quota predicate has not reset, output \texttt{WAIT\_QUOTA};
    \item else, if an active staleness predicate is stale, output \texttt{ASK\_CONFIRM};
    \item else, if an active cache predicate is expired, output \texttt{REFRESH};
    \item else, if an active job predicate is complete, output \texttt{RETRIEVE};
    \item else, if an active cache predicate is still valid, output \texttt{REUSE};
    \item otherwise output \texttt{WAIT}.
\end{enumerate}
This explicit priority policy is important because it defines what ``composition'' means in \ChronoState{}: the model must not only evaluate individual time predicates, but also apply a deterministic conflict-resolution order when several predicates are active.

\subsubsection{Families}

\ChronoState{} includes six families. Table~\ref{tab:families} summarizes the predicate and action structure for each family. The multi-constraint family composes multiple predicates under the priority policy in Sec.~\ref{sec:decision_policy}.

\begin{table}[t]
\centering
\caption{\ChronoState{} families. Each family requires elapsed time to be combined with visible symbolic state.}
\label{tab:families}
\small
\begin{tabular}{@{}L{0.18\columnwidth}L{0.41\columnwidth}L{0.27\columnwidth}@{}}
\toprule
\tabhead{Family} & \tabhead{Temporal predicate} & \tabhead{Typical actions}\\
\midrule
Cache & Is cached content still valid under its TTL? & \texttt{REUSE}/\texttt{REFRESH}\\
Job & Has a background job likely completed? & \texttt{WAIT}/\texttt{RETRIEVE}\\
Deadline & Is completion still feasible before deadline? & \texttt{WAIT}/\texttt{DEADLINE\_}\allowbreak\texttt{MISSED}\\
Quota & Has the quota window reset? & \texttt{WAIT\_}\allowbreak\texttt{QUOTA}/\texttt{RETRIEVE}\\
Staleness & Is the session too stale to continue directly? & \texttt{REUSE}/\texttt{ASK\_}\allowbreak\texttt{CONFIRM}\\
Multi & Priority composition of several active predicates & Mixed\\
\bottomrule
\end{tabular}
\end{table}

\subsubsection{Splits}

Per seed, \ChronoState{} contains 12,000 train examples, 2,000 validation examples, 4,000 standard-test examples, and 2,000 examples for each held-out split: template, duration, composition, and family. Across three seeds, this yields 78,000 raw records.

The train, validation, standard-test, held-out-template, and held-out-duration splits exclude pure quota. Quota appears during training only inside multi-constraint examples. The held-out-family split is pure quota. The held-out-composition split is pure multi with three or four active constraints.

Training elapsed-time values are
\begin{equation}
\{10,30,300,1800,7200,43200,86400,604800\}\ \mathrm{s},
\end{equation}
while held-out-duration values are
\begin{equation}
\{60,900,3600,10800,21600,172800,345600\}\ \mathrm{s}.
\end{equation}
The duration split therefore tests interpolation across unseen discrete durations in the same broad temporal range, not arbitrary extrapolation outside the benchmark's time support.

\subsection{Experimental setup}

This subsection specifies the frozen backbone, trainable surface, training protocol, evaluation conditions and controls, and scoring metrics used to report the results that follow.

\subsubsection{Model and trainable surface}

The reported \ChronoState{} experiment uses Qwen2.5-3B-Instruct as the frozen bf16 base model~\cite{qwen2025}. The model has 35 chrono injection sites and 145 LoRA modules. LoRA rank is 8 on \texttt{q\_proj}, \texttt{k\_proj}, \texttt{v\_proj}, \texttt{o\_proj}, and \texttt{lm\_head}. The same 31-dimensional chrono encoding is used throughout. The trainable \CI{} surface contains 9,577,472 of 3,095,516,160 parameters, or 0.309\%.

Because LoRA is applied to attention projections and the language-model head, the method should be interpreted as a frozen-backbone model with a small trained conditioning/action surface. It is not a claim that a completely frozen model, with no trained adapters, can solve \ChronoState{} from the hidden scalar alone.

Training uses 8,000 steps, learning rate $10^{-4}$, AdamW, weight decay 0.01, chunk length 512, forced-choice A/B/C/D next-token softmax loss, and log-probability scoring. Checkpoints are selected by validation balanced accuracy: seed 0 at step 6000, seed 1 at step 8000, and seed 2 at step 7000. Detailed reproducibility metadata, including run identifiers and checkpoint filenames, are provided in the artifact package rather than the main paper body.

\subsubsection{Conditions and controls}

\ChronoState{} evaluates six main conditions:
\begin{itemize}
    \item \emph{Hidden time}: true $\Tau$ supplied only through the hidden chrono channel.
    \item \emph{No-time control}: hidden $\Tau$ set to zero and no textual timestamp.
    \item \emph{Shuffled hidden time}: hidden $\Tau$ replaced by another example's $\Tau$ while the original label is kept.
    \item \emph{Prompt timestamp}: elapsed time written into the prompt while hidden $\Tau$ is zeroed.
    \item \emph{Both agree}: visible timestamp and hidden scalar agree.
    \item \emph{Conflict}: visible timestamp is wrong while hidden scalar is true.
\end{itemize}

The row labeled \emph{\CI{} model, timestamp text only; hidden $\Tau=0$} in Table~\ref{tab:chronostate_main} is an out-of-channel diagnostic: it asks whether the hidden-time trained \CI{} model can recover when elapsed time is moved into text and the hidden channel is zeroed. It is not the fair prompt baseline. The fair prompt-based comparison is \emph{Prompt+LoRA timestamp}, which is trained on textual timestamps. The LoRA-only, no-chrono-channel baseline removes the chrono path while retaining an otherwise comparable LoRA action surface. Vanilla no-time and vanilla prompt-timestamp baselines use the base model without the \ChronoState{} trainable surface.

\subsubsection{Metrics}

The primary metrics are action accuracy and balanced accuracy. Balanced accuracy is computed by averaging per-action recall over action labels that appear in the evaluated split. Because the selected action deterministically determines the updated temporal state under the oracle policy, state accuracy is equal to action accuracy in the reported main tables; it is included only as a compatibility metric.

Conflict examples report scalar-follow and prompt-follow. Let $\pi(s_i,\Tau_i)$ be the action implied by true hidden time, and let $\pi(s_i,\Tau_i^{\mathrm{text}})$ be the action implied by the visible timestamp. Scalar-follow is the fraction of conflict examples for which the prediction equals $\pi(s_i,\Tau_i)$; prompt-follow is the fraction for which the prediction equals $\pi(s_i,\Tau_i^{\mathrm{text}})$.

Shuffled-time analysis reports wrong-state consistency. Let $\tilde{\Tau}_i$ be the shuffled hidden time for example $i$, let $\hat{a}_i$ be the predicted action, let $a_i=\pi(s_i,\Tau_i)$ be the original gold action, and let $\tilde{a}_i=\pi(s_i,\tilde{\Tau}_i)$ be the action implied by the shuffled hidden time. With $C_i$ denoting the set of four choice actions, we compute
\begin{equation}
\WSC=\frac{1}{|\mathcal{I}|}\sum_{i\in\mathcal{I}}\mathbf{1}[\hat{a}_i=\tilde{a}_i],
\end{equation}
where
\begin{equation}
\mathcal{I}=\{i:\tilde{a}_i\neq a_i\ \mathrm{and}\ \tilde{a}_i\in C_i\}.
\end{equation}
This definition avoids counting cases where the shuffled-time-implied action is identical to the original label or is not available among the sampled answer choices.

%% file: sections/results_analysis.tex
\section{Results and Analysis}

This section reports the empirical evaluation of hidden-time chronometric injection on \ChronoState{}. We first present the main supervised results against no-time, shuffled-time, prompt-timestamp, LoRA-only, and vanilla baselines. We then analyze causal dependence through shuffled-time and conflict diagnostics, measure generalization on held-out template, duration, composition, and pure quota-family splits, and compare against linear diagnostic baselines that clarify what the benchmark measures. A short discussion interprets what these results support and what they do not.

\subsection{Hidden time composes with symbolic state under supervision}

Table~\ref{tab:chronostate_main} reports the main condition results. Hidden-time \CI{} reaches $\mathbf{0.9305\pm0.0134}$ accuracy and $\mathbf{0.9410\pm0.0103}$ balanced accuracy on the standard test. Both-agree performance is similar. The no-time control drops to $0.5511\pm0.0042$, and the shuffled-time control drops further to $0.3323\pm0.0097$. These controls show that the trained policy relies on the hidden scalar.

However, prompt+LoRA timestamp training is stronger, reaching $\mathbf{0.9893\pm0.0052}$ accuracy and $\mathbf{0.9920\pm0.0039}$ balanced accuracy. \ChronoState{} therefore should not be framed as evidence that hidden residual conditioning is better than prompt timestamps on supervised temporal-state tasks. The more defensible result is that hidden elapsed time is a viable, separately controllable conditioning interface.

\begin{table}[t]
\centering
\caption{\ChronoState{} main results. Accuracy, balanced accuracy, and state accuracy are reported as mean $\pm$ sample standard deviation over seeds 0, 1, and 2. The prompt+LoRA row is the fair text-time trained baseline and outperforms hidden-time \CI{} on this supervised benchmark.}
\label{tab:chronostate_main}
\small
\begin{tabular}{@{}p{0.40\textwidth}ccc@{}}
\toprule
\tabhead{Model / condition} & \tabhead{Accuracy} & \tabhead{Balanced acc.} & \tabhead{State acc.}\\
\midrule
\CI{} hidden time & $\mathbf{0.9305\pm0.0134}$ & $\mathbf{0.9410\pm0.0103}$ & $\mathbf{0.9305\pm0.0134}$ \\
\CI{} both agree & $0.9313\pm0.0072$ & $0.9403\pm0.0043$ & $0.9313\pm0.0072$ \\
\CI{} no-time control & $0.5511\pm0.0042$ & $0.5238\pm0.0122$ & $0.5511\pm0.0042$ \\
\CI{} model, timestamp text only; hidden $\Tau=0$ & $0.5428\pm0.0127$ & $0.5115\pm0.0230$ & $0.5428\pm0.0127$ \\
\CI{} shuffled hidden time & $0.3323\pm0.0097$ & $0.3557\pm0.0195$ & $0.3323\pm0.0097$ \\
LoRA-only, no chrono channel & $0.5513\pm0.0095$ & $0.5579\pm0.0433$ & $0.5513\pm0.0095$ \\
Prompt+LoRA timestamp & $\mathbf{0.9893\pm0.0052}$ & $\mathbf{0.9920\pm0.0039}$ & $\mathbf{0.9893\pm0.0052}$ \\
Vanilla no-time & $0.3731\pm0.0036$ & $0.3758\pm0.0082$ & $0.3731\pm0.0036$ \\
Vanilla prompt timestamp & $0.3738\pm0.0035$ & $0.3717\pm0.0123$ & $0.3738\pm0.0035$ \\
\bottomrule
\end{tabular}
\end{table}

\subsection{Shuffled-time controls show causal dependence}

The clearest causal diagnostic is shuffled hidden time. Across seeds, hidden-time accuracy is 0.9378, 0.9388, and 0.9150, while shuffled-time accuracy falls to 0.3243, 0.3295, and 0.3430. Wrong-state consistency is 0.9115, 0.9230, and 0.8835 (Table~\ref{tab:chronostate_shuffle}). Thus, when the injected scalar is wrong, the model tends to recompute the action from the wrong scalar rather than ignore it.

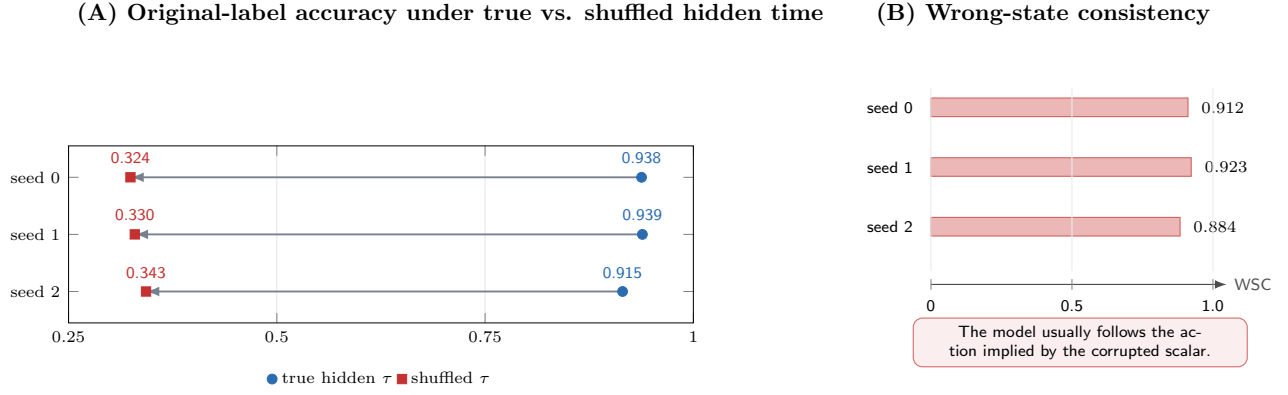
\begin{figure}[t]
\centering
\resizebox{0.985\textwidth}{!}{%
\begin{tikzpicture}[font=\sffamily\scriptsize,>=Latex,line width=0.48pt]
\node[anchor=west,font=\bfseries] at (0,4.65) {(A) Original-label accuracy under true vs. shuffled hidden time};
\begin{axis}[
    at={(0,0)},
    anchor=south west,
    width=11.0cm,
    height=4.25cm,
    xmin=0.25,xmax=1.00,
    ymin=-0.55,ymax=2.55,
    xlabel={},
    ytick={0,1,2},
    yticklabels={seed 2, seed 1, seed 0},
    xtick={0.25,0.50,0.75,1.00},
    xmajorgrids=true,
    grid style={black!10},
    tick label style={font=\scriptsize},
    label style={font=\scriptsize},
    legend style={font=\scriptsize, at={(0.5,-0.22)}, anchor=north, legend columns=2, draw=none},
]
\addplot[only marks, mark=*, mark size=2.1pt, ciBlue] coordinates {(0.9150,0) (0.9388,1) (0.9378,2)};
\addlegendentry{true hidden $\Tau$}
\addplot[only marks, mark=square*, mark size=2.0pt, ciRed] coordinates {(0.3430,0) (0.3295,1) (0.3243,2)};
\addlegendentry{shuffled $\Tau$}
\draw[ciGray!75, thick, -{Latex[length=2mm]}] (axis cs:0.9150,0) -- (axis cs:0.3430,0);
\draw[ciGray!75, thick, -{Latex[length=2mm]}] (axis cs:0.9388,1) -- (axis cs:0.3295,1);
\draw[ciGray!75, thick, -{Latex[length=2mm]}] (axis cs:0.9378,2) -- (axis cs:0.3243,2);
\node[anchor=south, text=ciBlue] at (axis cs:0.9150,0.10) {0.915};
\node[anchor=south, text=ciBlue] at (axis cs:0.9388,1.10) {0.939};
\node[anchor=south, text=ciBlue] at (axis cs:0.9378,2.10) {0.938};
\node[anchor=south, text=ciRed] at (axis cs:0.3430,0.10) {0.343};
\node[anchor=south, text=ciRed] at (axis cs:0.3295,1.10) {0.330};
\node[anchor=south, text=ciRed] at (axis cs:0.3243,2.10) {0.324};
\end{axis}

\node[anchor=west,font=\bfseries] at (12.05,4.65) {(B) Wrong-state consistency};
\def\wscw{4.25}
\foreach \name/\val/\y in {seed 0/0.9115/3.25,seed 1/0.9230/2.35,seed 2/0.8835/1.45}{
  \node[anchor=east] at (12.85,\y) {\name};
  \pgfmathsetmacro{\len}{\val*\wscw}
  \draw[fill=ciRed!35,draw=ciRed!75] (13.00,\y-0.14) rectangle ++(\len cm,0.28cm);
  \node[anchor=west] at (13.08+\len,\y) {\pgfmathprintnumber[fixed,precision=3]{\val}};
}
\draw[->,black!70] (13.00,0.58) -- (17.45,0.58) node[anchor=west] {WSC};
\foreach \x/\lab in {13.00/0,15.125/0.5,17.25/1.0}{
  \draw[black!45] (\x,0.50) -- (\x,0.66);
  \node[anchor=north] at (\x,0.46) {\lab};
  \draw[black!8] (\x,0.78) -- (\x,3.55);
}
\node[draw=ciRed!70, fill=ciRed!8, rounded corners=1.4mm, align=center, text width=4.8cm, anchor=north] at (15.25,0.08) {The model usually follows the action implied by the corrupted scalar.};
\end{tikzpicture}%
}
\caption{Shuffled-time intervention. Replacing the true hidden elapsed time with another example's scalar lowers original-label accuracy for every seed. High wrong-state consistency indicates that predictions usually match the action implied by the corrupted scalar when that action is one of the four choices.}
\label{fig:shuffle_intervention}
\end{figure}

This supports a precise causal claim: predictions depend causally on the injected scalar within the trained \ChronoState{} distribution. It does not by itself prove that the model has learned a fully abstract temporal rule independent of the synthetic generator distribution. Fig.~\ref{fig:shuffle_intervention} shows the seed-level paired intervention.

\begin{table}[t]
\centering
\caption{\ChronoState{} shuffled-time causal analysis. High wrong-state consistency means predictions agree with the action implied by the shuffled hidden time rather than the original label, on examples where that action is available among the choices.}
\label{tab:chronostate_shuffle}
\small
\begin{tabular}{@{}ccccc@{}}
\toprule
\tabhead{Seed} & \tabhead{CI} & \tabhead{Shuffled} & \tabhead{$\Delta$} & \tabhead{WSC}\\
\midrule
0 & 0.9378 & 0.3243 & 0.6135 & 0.9115 \\
1 & 0.9388 & 0.3295 & 0.6093 & 0.9230 \\
2 & 0.9150 & 0.3430 & 0.5720 & 0.8835 \\
\bottomrule
\end{tabular}
\end{table}

Conflict examples provide a complementary diagnostic. When prompt text and hidden time disagree, scalar-follow is $0.9065\pm0.0130$ and prompt-follow is $0.3323\pm0.0072$. This shows hidden-channel dominance for this trained model, but not safe disagreement handling. A deployed system should detect and handle time-source conflicts rather than silently follow one source.

\subsection{Generalization is strong except for pure quota-family transfer}

Table~\ref{tab:chronostate_generalization} reports split-level and family-level results. Hidden-time \CI{} remains strong on held-out templates at $\mathbf{0.9357\pm0.0109}$, held-out durations at $\mathbf{0.8375\pm0.0055}$, and held-out multi-constraint compositions at $\mathbf{0.8652\pm0.0438}$. These results show that the model is not merely memorizing exact prompt forms, exact training durations, or only single-predicate examples.

\begin{figure}[t]
\centering
\resizebox{\textwidth}{!}{%
\begin{tikzpicture}[font=\sffamily\scriptsize,>=Latex,line width=0.48pt]
\def\chartw{9.05}
\newcommand{\splitbar}[6]{%
  \pgfmathsetmacro{\len}{#2*\chartw}%
  \pgfmathsetmacro{\err}{#3*\chartw}%
  \node[anchor=east, text width=4.05cm, align=right] at (-0.20,#4+0.16) {#1};%
  \draw[fill=#5!62, draw=#5!92] (0,#4) rectangle (\len,#4+0.32);%
  \draw[black!65] (\len-\err,#4+0.16) -- (\len+\err,#4+0.16);%
  \draw[black!65] (\len-\err,#4+0.06) -- (\len-\err,#4+0.26);%
  \draw[black!65] (\len+\err,#4+0.06) -- (\len+\err,#4+0.26);%
  \node[anchor=west] at (\len+0.12,#4+0.16) {#6};%
}
\draw[->,black!70] (0,-3.30) -- (9.55,-3.30) node[anchor=west] {accuracy};
\foreach \x/\lab in {0/0,2.2625/0.25,4.525/0.50,6.7875/0.75,9.05/1.00}{
  \draw[black!45] (\x,-3.38) -- (\x,-3.22);
  \node[anchor=north] at (\x,-3.42) {\lab};
  \draw[black!8] (\x,0.62) -- (\x,-3.16);
}
\splitbar{Held-out template}{0.9357}{0.0109}{0.00}{ciBlue}{0.936}
\splitbar{Standard test}{0.9305}{0.0134}{-0.70}{ciBlue}{0.931}
\splitbar{Held-out composition}{0.8652}{0.0438}{-1.40}{ciBlue}{0.865}
\splitbar{Held-out duration}{0.8375}{0.0055}{-2.10}{ciBlue}{0.838}
\splitbar{Held-out family: quota}{0.5065}{0.0559}{-2.80}{ciRed}{0.507}
\node[draw=ciRed!70, fill=ciRed!8, rounded corners=1.4mm, align=center, text width=3.55cm, anchor=west] (callout) at (6.25,-2.78) {main limitation:\\pure quota-family transfer};
\draw[ciRed!80, thick, -{Latex[length=2mm]}] (callout.west) to[out=180,in=20] (4.65,-2.48);
\end{tikzpicture}%
}
\caption{Generalization and limitation summary for hidden-time \CI{}. Performance remains high on held-out templates, held-out durations, and held-out multi-constraint compositions, but drops sharply on the pure held-out quota-family split.}
\label{fig:generalization}
\end{figure}
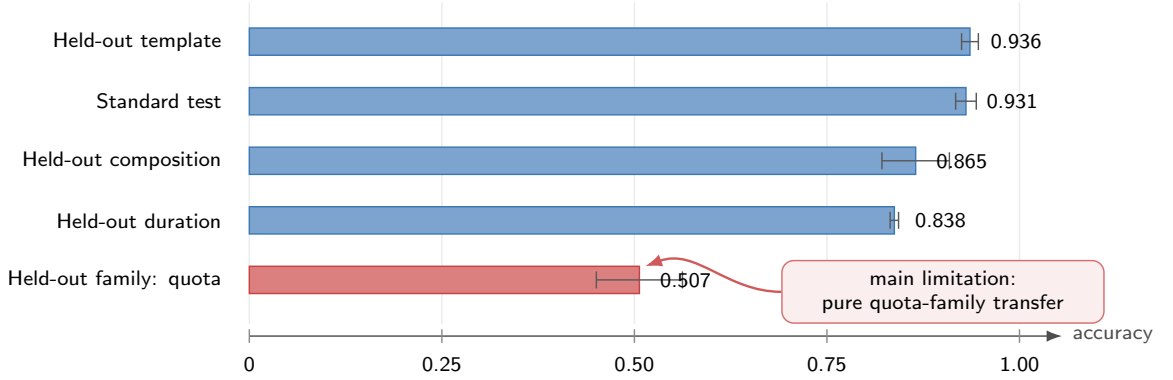

The held-out-family split is different. Pure quota examples are excluded from training, although quota appears inside multi-constraint examples. On the pure quota held-out-family split, \CI{} reaches only $\mathbf{0.5065\pm0.0559}$ accuracy and $0.5695\pm0.0496$ balanced accuracy. Many true \texttt{RETRIEVE} cases are predicted as \texttt{WAIT\_QUOTA}. This is the main generalization failure and limits any claim of broad predicate abstraction. Fig.~\ref{fig:generalization} emphasizes this scope boundary.

\begin{table}[t]
\centering
\caption{\ChronoState{} generalization and family results. The held-out-family split is pure quota; quota appears in training only inside multi-constraint examples.}
\label{tab:chronostate_generalization}
\small
\begin{tabular}{@{}lccc@{}}
\toprule
\tabhead{Split} & \tabhead{\CI{} acc.} & \tabhead{\CI{} bacc.} & \tabhead{LoRA-only acc.}\\
\midrule
Standard test & $\mathbf{0.9305\pm0.0134}$ & $\mathbf{0.9410\pm0.0103}$ & $0.5513\pm0.0095$ \\
Held-out template & $\mathbf{0.9357\pm0.0109}$ & $\mathbf{0.9473\pm0.0096}$ & $0.5468\pm0.0059$ \\
Held-out duration & $\mathbf{0.8375\pm0.0055}$ & $\mathbf{0.8577\pm0.0064}$ & $0.5650\pm0.0046$ \\
Held-out composition & $\mathbf{0.8652\pm0.0438}$ & $\mathbf{0.8006\pm0.0313}$ & $0.5052\pm0.0389$ \\
Held-out family, quota & $\mathbf{0.5065\pm0.0559}$ & $\mathbf{0.5695\pm0.0496}$ & $0.4238\pm0.0435$ \\
\bottomrule
\end{tabular}
\vspace{0.6em}

\small
\begin{tabular}{@{}lcc@{}}
\toprule
\tabhead{Family} & \tabhead{Acc.} & \tabhead{Balanced acc.}\\
\midrule
Cache & $0.9711\pm0.0189$ & $0.9711\pm0.0189$ \\
Deadline & $0.9589\pm0.0255$ & $0.9298\pm0.0438$ \\
Job & $0.9459\pm0.0173$ & $0.9459\pm0.0173$ \\
Staleness & $0.9550\pm0.0090$ & $0.9465\pm0.0191$ \\
Multi & $0.9028\pm0.0126$ & $0.9159\pm0.0063$ \\
Quota held-out & $0.5065\pm0.0559$ & $0.5695\pm0.0496$ \\
\bottomrule
\end{tabular}
\end{table}

\subsection{Linear diagnostics clarify what the benchmark measures}

Table~\ref{tab:linear_diagnostics} reports non-neural diagnostic baselines. A linear model over $\chi(\Tau)$ alone is weak on standard, template, and duration splits. State-only features are also insufficient. A linear model over $\chi(\Tau)$ plus engineered state features reaches high performance on standard, template, duration, and composition splits, and outperforms \CI{} on the held-out quota family. The rule oracle reaches 1.0000 on all splits.

These diagnostics clarify that \ChronoState{} is not intrinsically unsolvable. Explicit state+time features are enough. The learned hidden-time model captures much of this structure under the trained family mixture but does not abstract pure quota as well as the engineered diagnostic baseline.

\begin{table}[t]
\centering
\caption{\ChronoState{} linear diagnostic baselines. Values are accuracies. These are diagnostic ceilings or feature tests, not neural model baselines.}
\label{tab:linear_diagnostics}
\small
\begin{tabular}{@{}lccccc@{}}
\toprule
\tabhead{Baseline} & \tabhead{Std.} & \tabhead{Template} & \tabhead{Duration} & \tabhead{Composition} & \tabhead{Family}\\
\midrule
$\chi(\Tau)$ only & 0.3103 & 0.3163 & 0.2163 & 0.5073 & 0.3848 \\
State only & 0.5180 & 0.5142 & 0.5388 & 0.4580 & 0.1753 \\
$\chi(\Tau)$ + state & 0.9230 & 0.9265 & 0.8220 & 0.8138 & 0.7137 \\
Rule oracle & 1.0000 & 1.0000 & 1.0000 & 1.0000 & 1.0000 \\
\bottomrule
\end{tabular}
\end{table}

\subsection{Discussion}

We next interpret the empirical findings: what \ChronoState{} supports as an interface and benchmark result, what it does not support, how novelty should be scoped, and what the architectural and deployment implications are.

\subsubsection{What \ChronoState{} supports}

\ChronoState{} supports supervised hidden-time composition. The hidden scalar is not merely used as a label-family threshold: the model must combine elapsed time with symbolic state variables and choose among a shared action set. The no-time, LoRA-only, and shuffled-time controls show that the hidden scalar is necessary for high \ChronoState{} accuracy in the learned policy. The held-out-template, held-out-duration, and held-out-composition results show useful generalization beyond exact prompt forms, exact training durations, and simple single-predicate examples.

The strongest causal evidence is wrong-state consistency under shuffled hidden time. When the scalar is shuffled, predictions tend to agree with the action implied by the shuffled scalar. This indicates that the model recomputes the temporal state from the injected time, even when that time is wrong.

A useful way to interpret the contribution is to separate three claims. First, as an interface result, a hidden scalar can be injected and causally controlled. Second, as a benchmark result, \ChronoState{} operationalizes temporal-state composition through explicit rules over state and elapsed time. Third, as a capability result, this trained frozen-backbone model solves several supervised families and splits, but does not robustly transfer to the pure held-out quota family.

\subsubsection{What \ChronoState{} does not support}

The supported claim is intentionally narrow. \ChronoState{} shows that, under direct supervision, a frozen-backbone model with a small trainable surface can compose a system-supplied non-token elapsed-time scalar with visible symbolic state to select temporal actions, and that scalar interventions (zero, shuffle, conflict) can diagnose whether that channel is used. It does \emph{not} show autonomous time tracking: $\Tau$ is supplied externally, and a wrong value can drive a wrong action. It does \emph{not} show superiority over prompt timestamps: prompt+LoRA remains substantially stronger on this supervised benchmark. It does \emph{not} establish broad unseen-family transfer: pure held-out quota remains weak, suggesting incomplete abstraction of that predicate when it appears only inside multi-constraint training examples. The defensible scientific result is a controllable interface and evaluation protocol, not a claim of general temporal intelligence.

\subsubsection{How the novelty should be stated}

The novelty claim should remain combination-level rather than component-level. Time-aware LMs, continuous timestamp prompts, residual activation interventions, FiLM conditioning, adapters, and LoRA are all established ideas. The distinct proposal is the following controlled setup: a system-supplied elapsed-time scalar is hidden from the token stream, injected into a frozen-backbone LM through a small trained conditioning surface, composed with visible symbolic state, and evaluated with zero, shuffled, prompt-conflict, and held-out-composition controls. This is narrower than claiming a new general theory of temporal reasoning, but stronger than simply reporting that a timestamp feature helps a classifier.

\subsubsection{Implications}

The main implication is architectural and methodological. For systems where elapsed time should influence internal routing or action selection without being exposed as visible text, a residual-stream hidden-time channel is viable under direct supervision. For research, the channel makes time-source interventions easy: $\Tau$ can be zeroed, shuffled, or put into conflict with visible text.

The deployment implication is cautious. Hidden-time channels should be paired with validated time sources, validation, and conflict handling. A model that follows hidden time strongly may behave poorly if the scalar is stale, corrupted, or inconsistent with visible context. The result should be viewed as evidence for a controllable conditioning interface, not as a replacement for symbolic validation around time-sensitive decisions.

%% file: sections/conclusion.tex
\section{Conclusion and Future Work}

\ChronoState{} shows that hidden elapsed time can be composed with symbolic task state in a frozen-backbone language model under direct supervision. A chronometric-injection model with a small trainable surface reaches high accuracy on standard, template, duration, and composition splits, and causal controls show that correct hidden time is necessary for high accuracy. The same results also define the boundary: prompt+LoRA timestamps are stronger, wrong hidden time drives wrong predictions, and pure held-out quota-family transfer remains weak.

In narrow terms, we provide a benchmark and conditioning interface for testing whether a frozen-backbone LM can use a system-supplied non-token elapsed-time scalar as system state when selecting actions from symbolic temporal task descriptions. We do not claim that hidden time replaces code, beats prompt timestamps, or gives a model autonomous awareness of time.

Future work should move from deterministic synthetic tasks to external temporal-state environments, add stronger matched scalar and prefix baselines, evaluate larger and more diverse base models, add leave-one-family-out evaluations for every predicate family, and develop explicit conflict-detection policies for disagreement between visible time text and system-supplied hidden time. Such extensions (especially on real-time sensing, multi-domain decision support, and secure agent tooling) align with ongoing SPIE Defense~+~Security interest in trustworthy, time-aware autonomy and machine-assisted analysis.

\subsection{Limitations}

The benchmark is synthetic and deterministic. This is useful for controlled causal analysis but does not guarantee external validity. The model family is limited to Qwen2.5-3B-Instruct, with three seeds and no broad scale sweep. The time range is finite, and the reported held-out durations are interpolation-style held-outs rather than arbitrary extrapolation outside the benchmark's time support.

The strongest prompt-based baseline, prompt+LoRA, outperforms hidden-time \CI{}. Additional matched baselines remain useful, including soft-prefix tuning, a matched scalar-embedding input channel, chrono-only without LoRA, LoRA only on the language-model head, LoRA without the language-model head, different LoRA ranks, and injection at fewer layers. The time encoding is fixed rather than justified by ablation; raw scalar, log-only, sinusoid-only, learned scalar encodings, and Time2Vec-style learned features should be evaluated.

Conflict handling is underdeveloped. Scalar-follow is high, but safe abstention or disagreement detection is not demonstrated. The conflict test measures source dominance, not safe arbitration. A real system should validate its time source and specify what happens when visible and hidden time disagree. Future experiments should include explicit conflict labels such as \texttt{TIME\_CONFLICT}, \texttt{ASK\_CONFIRM}, or \texttt{USE\_TRUSTED\_CLOCK}.

The held-out quota-family failure is the largest substantive limitation. Future versions of \ChronoState{} should include leave-one-family-out splits for cache, job, deadline, quota, staleness, and multi families, plus diagnostic interventions to determine whether failures arise from action priors, predicate abstraction, or prompt-template mismatch.

Related-work overlap should also be handled carefully. Similar components already exist: timestamp-conditioned LMs, continuous time prompts, prefix/adapters, residual activation steering, and temporal-agent evaluations. The paper should not claim that time injection, continuous time features, or activation conditioning are new in isolation. The defensible novelty is the controlled combination of hidden elapsed scalar conditioning with symbolic temporal-state action selection and causal scalar interventions.

Reproducibility is strong at the artifact level but incomplete at the upstream-model level. The reproducibility package records seeds, checkpoint choices, software versions, scripts, and training details, but the exact upstream Qwen model revision was not recorded. Future releases should include a recorded upstream-model revision.

Looking ahead, evaluating hidden-time conditioning on operationally motivated tasks (track-currency checks, sensor-product staleness, and multi-constraint mission-support routing) would clarify transfer beyond synthetic predicates and is a natural fit for defense and security AI venues within the SPIE community.